\pdfoutput=1

\documentclass[11pt]{article}

\usepackage[]{EMNLP2022}

\usepackage{times}
\usepackage{latexsym}
\usepackage{times}
\usepackage{latexsym}
\usepackage{graphicx}
\usepackage{adjustbox}
\usepackage{booktabs}
\usepackage{todonotes}
\usepackage{multirow}
\usepackage{amsmath}
\usepackage[ruled,vlined]{algorithm2e}

\usepackage[T1]{fontenc}

\usepackage[utf8]{inputenc}

\usepackage{microtype}
\usepackage{amssymb}
\usepackage{pifont}
\newcommand{\cmark}{\ding{51}}%
\newcommand{\xmark}{\ding{55}}%
\title{Exploiting Rich Textual User-Product Context for Improving Sentiment Analysis}

 \author{
  \textbf{Chenyang Lyu}$^\dag$
  ~~~~ \textbf{Linyi Yang}$^\ddag$~~~~ 
 \textbf{Yue Zhang}$^\ddag$~~~~ \textbf{Yvette Graham}$^\P$~~~~ \textbf{Jennifer Foster}$^\dag$~~~~  \\
  $^\dag$ School of Computing, Dublin City University, Dublin, Ireland \\
  $^\ddag$ School of Engineering, Westlake University, China \\
  $^\P$ School of Computer Science and Statistics, Trinity College Dublin, Dublin, Ireland\\
  \texttt{chenyang.lyu2@mail.dcu.ie}, \texttt{ygraham@tcd.ie}, \texttt{jennifer.foster@dcu.ie} \\
  \texttt{{\{yanglinyi, zhangyue\}}@westlake.edu.cn}
}

\begin{document}
\maketitle

\begin{abstract}
User and product information associated with a review is useful for sentiment polarity prediction. Typical approaches incorporating such information focus on modeling users and products as implicitly learned representation vectors.
Most do not exploit the potential of historical reviews, or those that currently do require unnecessary modifications to model architecture or
do not make full use of 
user/product
associations.
The contribution of this work is twofold: i) a method to explicitly employ
historical reviews belonging to the same user/product to initialize  representations, 
and ii) efficient incorporation of textual associations between users and products via a user-product cross-context module.
Experiments on IMDb, Yelp-2013 and Yelp-2014 benchmarks show that our approach substantially outperforms previous state-of-the-art. Since we employ BERT-base as the encoder, we additionally provide experiments in which our approach performs well with SpanBERT and Longformer. Furthermore, experiments where the reviews of each user/product in the training data are downsampled demonstrate the effectiveness of our approach under a low-resource setting. 



\end{abstract}

\section{Introduction}


It has been repeatedly shown that the user and product information associated with reviews is helpful for sentiment polarity prediction~\cite{tang-etal-2015-user-product,tang-etal-2015-user-product,chen-etal-2016-neural-sentiment,ma-etal-2017-cascading}. Just as the same user is expected to have consistent narrative style and vocabulary, the reviews belonging to the same product are expected to exhibit similar vocabulary for specific terms. Most previous work models user and product identities as representation vectors which are implicitly learned during the training process and only focus on the interactions between the user or product and the review text~\cite{dou-2017-capturing,long-etal-2018-dual,amplayo-2019-rethinking,zhang-etal-2021-ma-bert}. This brings with it two major shortcomings: i) the associations between users and products are not fully exploited, and, ii) the text of historical reviews is not used. 

To tackle the first shortcoming, \newcite{amplayo-etal-2018-cold} propose to incorporate similar user and product representations for review sentiment classification. However, their approach requires a complex selective gating mechanism while 
ignoring the associations between users and products. To tackle the second shortcoming, \newcite{lyu-etal-2020-improving} propose to explicitly use historical reviews in the training process. However, their approach needs to incrementally store review representations during the training process, which results in a more complex model architecture.

\begin{figure}
    \centering
    \includegraphics[width=\linewidth]{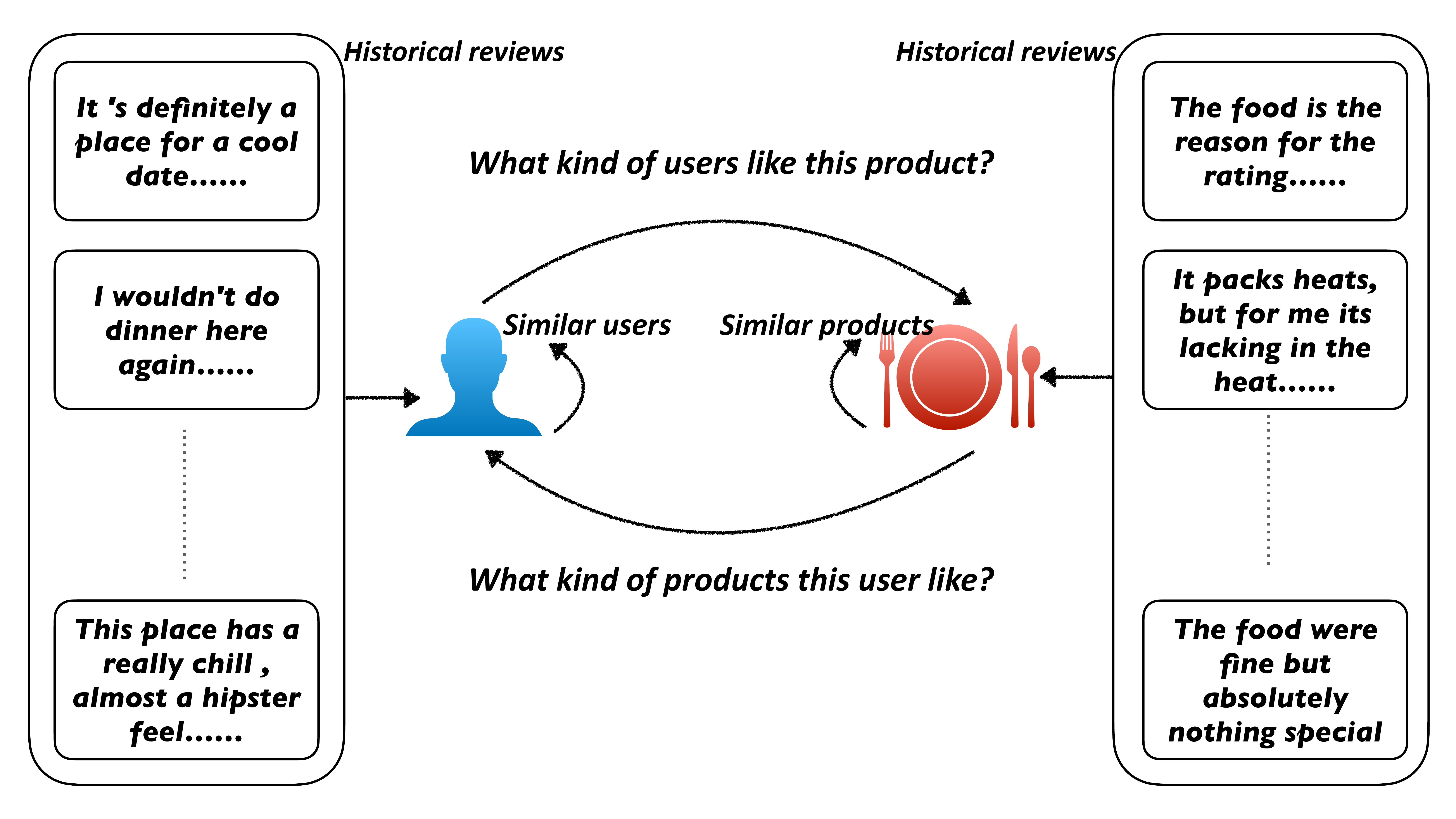}
    \caption{Our proposed idea of representing users and products with their historical reviews, which can directly inform user and product preferences, and incorporating the associations between users and products.}
    \label{fig:idea_sketch}
\end{figure}

As shown in Figure~\ref{fig:idea_sketch}, we propose two simple strategies to address the aforementioned issues. Firstly, we use pre-trained language models (PLMs) to pre-compute the representations of all historical reviews belonging to the same user/product. Historical review representations are then used to initialize user/product representations by average pooling over all tokens before again average pooling over all reviews. 
This allows historical review text to inform the user and product preference, which we believe is potentially more advantageous than implicitly learned user/product representations. In addition, time and memory costs are minimized 
since the representations of historical reviews are average pooled
and the pre-computation is one-time. 

Secondly, 
we propose a user-product cross-context module, which cooperates with historical representations of users and products to gather sentiment polarity information from the reviews of other users/products. This module interacts on four dimensions: user-to-user, product-to-product, user-to-product and product-to-user. The former two are used to obtain similar user (product) information, which is useful to model user (product) preference especially when a user (product) has limited reviews. The latter two are used to model the product preference of the user (what kind of products do they like and what kind of ratings would they give to similar products?) and user preference associated with a product (what kinds of users like such products and what kinds of ratings would they give to this product?). 


We apply our approach to various English PLMs and test on three benchmark English datasets -- IMDb, Yelp-2013, Yelp-2014. We find that our approach yields consistent improvements across PLMs and achieves substantial improvements over previous state-of-the-art models. 
We also show the superior performance of our approach when the number of reviews for each user is limited.

Our contributions are two effective, cooperative strategies for improving sentiment analysis with user and product information:
\begin{enumerate}
\setlength\itemsep{-0.5em}
\item initializing user and product representations using their historical reviews 
\item a user-product cross-context module which cooperates with Contribution 1 to efficiently incorporate textual associations between users and products from a larger context.
\end{enumerate}

\section{Related Work}
Neural models have been widely used in sentiment classification \cite{socher-etal-2013-recursive,kim-2014-convolutional,ibm_cnn,tang-etal-2015-document,wang2016attention}. Most existing works focus purely on text. However, \citep{tang-etal-2015-user-product} point out the importance of incorporating user and product information for review sentiment classification, where such information is available. Subsequently various methods have been explored to inject user and product information~\cite{tang-etal-2015-user-product,chen-etal-2016-neural-sentiment,ma-etal-2017-cascading,dou-2017-capturing,long-etal-2018-dual,amplayo-etal-2018-cold,amplayo-2019-rethinking} based on CNN and LSTM.
After the emergence of PLMs such as BERT~\cite{bert}, the performance of this task has significantly improved. 
\newcite{zhang-etal-2021-ma-bert}, for example, proposed a multi-attribute encoder using bilinear projections between attributes and texts on top of BERT~\cite{bert} to make better use of attribute (user and product) information.

How to utilize the user and product information has been investigated comprehensively among aforementioned studies. 
However, most focus on modeling user and product as randomly initialized embedding vectors implicitly learned in training. \newcite{lyu-etal-2020-improving} demonstrate that explicitly using historical reviews can substantially improve the performance of a sentiment classification model by incrementally storing their representations during the training process.  However, the magnitude of the user and product matrix is difficult to control when the number of reviews grow very large. As well as that, the associations among users and products are not fully utilized. Although~\newcite{amplayo-etal-2018-cold} propose to make use of information about similar users and similar products, the information about users associated with the current product is not taken into consideration.
\section{Methodology}
\begin{figure*}
    \centering
    \includegraphics[scale=0.185]{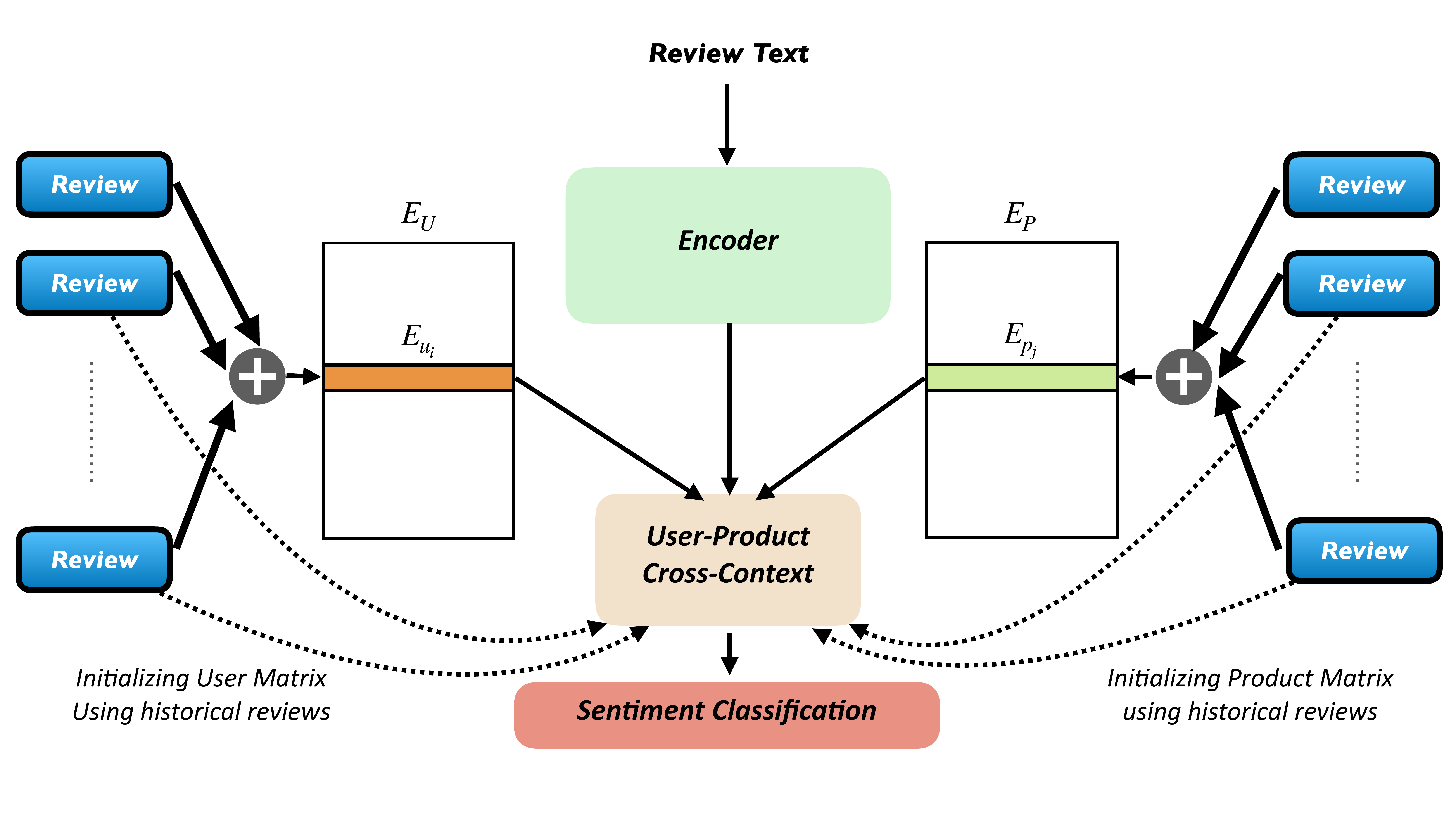}
    \caption{Our model architecture. We initialize user representation matrix $E_{U}$ and product representation matrix $E_{P}$. The user vector $E_{u_{i}}$ and product vector $E_{p_{j}}$ are fed into user-product cross-context module with document representation $H_{D}$. The dashed lines indicate the direct interactions of historical reviews in the cross-context module.}
    \label{fig:model_architecture}
\end{figure*}

An overview of our approach is shown in Figure~\ref{fig:model_architecture}. We firstly feed the review text, $D$, into a PLM encoder to obtain its representation, $H_{D}$. $H_{D}$ is then  fed into a user-product cross-context module consisting of multiple attention functions together with the corresponding user embedding vector, $u$ and product embedding vector, $p$. The output of the user-product module is concatenated with $H_{D}$ and fed into a linear classification layer to obtain the distribution over all sentiment labels. The architecture design is novel in two ways: 1) the user and product embedding matrix is initialized using representations of historical reviews of the corresponding users/products, 2) a user-product cross-context module that cooperates with 1) to incorporate textual associations between users and products. 
\subsection{Incorporating Textual Information of Historical Reviews}
For the purpose of making use of the textual information of historical reviews, we initialize all user and product embedding vectors using the representations of their historical reviews. Specifically, assume that we have a set of users $U=\{u_{1},......,u_{N}\}$ and products $P=\{p_{1},......,p_{M}\}$. Each user $u_{i}$ and product $p_{j}$ have their corresponding historical reviews: $u_{i}=\{D_{1}^{u_{i}}, ......, D_{n_{i}}^{u_{i}}\}$ and $p_{j}=\{D_{1}^{p_{j}}, ......, D_{m_{j}}^{p_{j}}\}$. 

For a certain user $u_{i}$, we firstly feed $D_{1}^{u_{i}}$ into the transformer encoder to obtain its representation $H_{D_{1}}^{u_{i}} \in \mathbf{R}^{L\times h}$, then we average $H_{D_{1}}^{u_{i}}$ along its first dimension:
\begin{equation}
    \bar{H}_{D_{1}}^{u_{i}} = \frac{\sum{H_{D_{1}}^{u_{i}}}}{T_{D_{1}}^{u_{i}}}
\end{equation}
where $\bar{H}_{D_{1}}^{u_{i}}\in \mathbf{R}^{1\times h}$, $L$ is the maximum sequence length, $h$ is the hidden size of the transformer encoder, $T_{D_{1}}^{u_{i}}$ is the total number of tokens in $D_{1}^{u_{i}}$ excluding special tokens. Therefore, we simply sum the representations of all tokens in $D_{1}^{u_{i}}$ and then average it to obtain a document vector $\bar{H}_{D_{1}}^{u_{i}}$. The same procedure is used to generate the document vectors of all documents in $u_{i}=\{D_{1}^{u_{i}}, ......, D_{n_{i}}^{u_{i}}\}$. Finally, we obtain the representation of $u_{i}$ by:
\begin{equation}
    E_{u_{i}} = \frac{\sum_{k=1}^{n_{i}}{\bar{H}_{D_{k}}^{u_{i}}}}{n_{i}}
\end{equation}
where $E_{u_{i}} \in \mathbf{R}^{1\times h}$ is the initial representation of user $u_{i}$. The same process is applied to generate the representations of all the other users as well as all products. Finally, we have $E_{U} \in \mathbf{R}^{N\times h}$ and $E_{P} \in \mathbf{R}^{M\times h}$ as the user and product embedding matrix respectively. Moreover, in order to control the magnitude of $E_{U}$, $E_{P}$ 
to prevent it from being too large or too small, 
we propose scaling heuristics, $E_{U}$ and $E_{P}$:
\begin{equation}
        \hat{E_{U}} = f_{U}E_{U},
        f_{U} = \frac{F\_Norm(E)}{F\_Norm(E_{U})}
\label{user_scaling}
\end{equation}
\begin{equation}
        \hat{E_{P}} = f_{P}E_{P}, 
        f_{P} = \frac{F\_Norm(E)}{F\_Norm(E_{P})}
\label{product_scaling}
\end{equation}
where $F\_Norm$ is Frobenius norm, $E$ is a normal matrix in which the elements $E_{i,j}$ is drawn from a normal distribution $\mathcal{N}(0,1)$. 
\subsection{User-Product Information Integration}
Having enriched user and product representations with historical reviews, we 
propose a user-product cross-context module for the purpose of 
garnering
sentiment clues from textual associations between users and products.  In 
this
module, we adopt four attention operations: \textit{user-to-user, product-to-product, user-to-product} and \textit{product-to-user}. We use \textit{Multi-Head-Attention}~\cite{transformer} in four attention operations. Specifically, for \textit{Multi-Head-Attention(Q,K,V)}, we use the user representation $E_{u_{i}}$ or product representation $E_{p_{j}}$ as \textit{Q} and the user matrix $E_{U}$ and product matrix $E_{P}$ as \textit{K} and \textit{V}. It is important to note that, before using $E_{u_{i}}$ and $E_{p_{j}}$, we fuse the document information $H_{cls} \in \mathbf{R}^{1\times h}$, the representation of the \textit{[CLS]} token, into them as follows:
\begin{equation}
    E_{u_{i}} = g_{u}(E_{u_{i}}, H_{cls}), E_{p_{j}} = g_{p}(E_{p_{j}}, H_{cls}), 
\end{equation}
where $g_{u}$ and $g_{p}$ represent two linear layers combining $E_{u_{i}}/E_{p_{j}}$ and $H_{cls}$.

\paragraph{User-to-User Attention} We use $E_{u_{i}}$ as the query and $E_{U}$ as the keys and values to gather information from similar users:
\begin{equation}
    E_{u_{i}}^{uu} = Attn_{uu}(E_{u_{i}},E_{U}, E_{U})
\end{equation}
\paragraph{Product-to-Product Attention} We use $E_{p_{j}}$ as the query and $E_{P}$ as the keys and values to gather information from similar products:
\begin{equation}
    E_{p_{j}}^{pp} = Attn_{pp}(E_{p_{j}},E_{P}, E_{P})
\end{equation}
\paragraph{User-to-Product Attention} We use $E_{u_{i}}$ as the query and $E_{P}$ as the keys and values to gather information from products associated with $u_{i}$:
\begin{equation}
    E_{u_{i}}^{up} = Attn_{up}(E_{u_{i}},E_{P}, E_{P})
\end{equation}
\paragraph{Product-to-User Attention} We use $E_{p_{j}}$ as the query and $E_{U}$ as the keys and values to gather information from users associated with $p_{j}$:
\begin{equation}
    E_{p_{j}}^{pu} = Attn_{pu}(E_{p_{j}},E_{U}, E_{U})
\end{equation}
%
We also employ two \textit{Multi-head Attention} between $E_{u_{i}}$/$E_{p_{j}}$~(query) and $H_{D}$~(key and value). The corresponding outputs are $E_{u_{i}}^{D}$ and $E_{p_{j}}^{D}$.
We then combine the output of the user-product cross-context module and $H_{cls}$ to form the final representations. In $Attn_{uu}$ and $Attn_{pp}$, we add attention masks to prevent $E_{u_{i}}$ and $E_{p_{j}}$ from attending to themselves. Thus we also incorporate $E_{u_{i}}$ and $E_{p_{j}}$ as their \textit{self-attentive} representations:
\begin{equation}
\begin{split}
        H_{d}  = & g(E_{u_{i}}^{uu}, E_{p_{j}}^{pp}, E_{u_{i}}^{up}, E_{p_{j}}^{pu}, E_{u_{i}}^{D}, E_{p_{j}}^{D}, \\
        & E_{u_{i}}, E_{p_{j}},H_{cls})
\end{split}
\end{equation}
$H_{d}$ is fed into the classification layer to obtain the sentiment label distribution.

We use \textit{Cross-Entropy} to calculate the loss between our model predictions and the gold labels.


\section{Experiments}



\begin{table}[!htb]
  \centering
  \small
      \begin{tabular}{lrrccccc}
\toprule
    Datasets   & Train & Dev  & Test & Words/Doc\\
    \midrule
    IMDB       & 67,426 & 8,381 & 9,112 & 394.6  \\
    Yelp-2013  & 62,522 & 7,773 & 8,671 & 189.3  \\
    Yelp-2014  & 183,019 & 22,745 & 25,399 & 196.9  \\
    \bottomrule
\end{tabular}%
      \caption{Number of documents in per split and average doc length of IMDb, Yelp-2013 and Yelp-2014.}
       \label{data_statistics_train}
\end{table}

\begin{table}[!htb]
  \centering
  \resizebox{\linewidth}{!}{
      \begin{tabular}{lcccc}
\toprule
    Datasets   & Users & Products  & Docs/User & Docs/Product\\
    \midrule
    IMDB      &   1,310 & 1,635 & 64.82 & 51.94 \\
    Yelp-2013 &   1,631 & 1,633 & 48.42 & 48.36 \\
    Yelp-2014 &   4,818 & 4,194 & 47.97 & 55.11 \\
    \bottomrule
\end{tabular}%
            }
      \caption{Number of users and products with average amount of documents for each user and product in IMDb, Yelp-2013 and Yelp-2014.}
\label{data_statistics_user_product}

\end{table}


\begin{table*}[!htb]
  \centering
  \small
  \begin{tabular}{lcccccccccccccccc}
    \toprule
     && \multicolumn{2}{c}{IMDB} && \multicolumn{2}{c}{Yelp-2013}  && \multicolumn{2}{c}{Yelp-2014} \\ 
    
     && Acc. (\%) &  RMSE &&  Acc. (\%) &  RMSE &&  Acc. (\%) & RMSE  \\
     
    \midrule

        Vanilla BERT-base Attention && 55.4 & 1.129 && 69.1 & 0.617 && 70.7  & 0.610 \\

        \hspace{0.1cm} + Our approach && \textbf{59.7} & \textbf{1.006} && \textbf{70.7}  & \textbf{0.589} && \textbf{72.4}  & \textbf{0.559} \\
        
        
        Vanilla BERT-large Attention && 55.7 & 1.070 && 69.9  & 0.590 && 71.3  & 0.579\\
        \hspace{0.1cm} + Our approach && \textbf{60.3} & \textbf{0.977} && \textbf{71.8} & \textbf{0.568} && \textbf{72.3}  & \textbf{0.567} \\
        
        
        Vanilla SpanBERT-base Attention && 56.6 & 1.055 && 70.2  & 0.589 && 71.3  & 0.571 \\
        \hspace{0.1cm} + Our approach && \textbf{60.2} & \textbf{1.026} && \textbf{71.5} & \textbf{0.578} && \textbf{72.6}  & \textbf{0.562} \\
        
        
        Vanilla SpanBERT-large Attention && 57.6 & 1.009 && 71.6  & 0.563 && 72.5  & 0.556 \\
        \hspace{0.1cm} + Our approach && \textbf{61.0} & \textbf{0.947} && \textbf{72.7} & \textbf{0.552} && \textbf{73.7} & \textbf{0.543} \\
        
        

        Vanilla Longformer-base Attention && 56.7 & 1.019 && 71.0  & 0.573 && 72.5  & 0.554 \\
        \hspace{0.1cm} + Our approach && \textbf{59.6} & \textbf{0.990} && \textbf{72.6}  & \textbf{0.558} && \textbf{73.3}  & \textbf{0.548} \\
        
        Vanilla Longformer-large Attention && 57.0 & 0.967 && 70.7  & 0.571 && 72.2  & 0.555 \\
        \hspace{0.1cm} + Our approach && \textbf{61.8} & \textbf{0.931} && \textbf{73.5}  & \textbf{0.540} && \textbf{74.3}  & \textbf{0.529} \\

    \bottomrule
    \end{tabular}%
    \caption{Results of our approach on various PLMs on the dev sets of IMDb, Yelp-2013 and Yelp-2014. We show the results of the baseline vanilla attention model for each PLM as well as the results of the same PLM with our proposed approach. We report the average of five runs with two metrics, Accuracy~($\uparrow$) and RMSE~($\downarrow$)}
  \label{tbl-03-more-models-results}%
\end{table*}

\subsection{Datasets}
Our experiments are conducted on three benchmark English document-level sentiment analysis datasets: IMDb, Yelp-13 and Yelp-14~\cite{tang-etal-2015-user-product}. Statistics of the three datasets are shown in Table~\ref{data_statistics_train}. 
The IMDb dataset has the longest documents with an average length of approximately 395 words. All three are fine-grained sentiment analysis datasets: Yelp-2013 and Yelp-2014 have 5 classes, IMDb has 10 classes. 
Each review is accompanied by its corresponding anonymized user ID and product ID. 
The average number of reviews for each user/product is shown in Table~\ref{data_statistics_user_product}.

\subsection{Experimental Setup}
The pre-trained language models we employed in experiments are BERT-base-uncased, BERT-large-uncased~\cite{bert}, SpanBERT-base, SpanBERT-large~\cite{joshi2020spanbert} and Longformer~\cite{beltagy2020longformer}. We use the implementations from Huggingface~\cite{Wolf2019HuggingFacesTS}. The hyperparameters are empirically selected based on the performance on the dev set. We adopt a early stopping strategy where we stop training when the performance on dev set decreases. The maximum sequence is set to 512 for all models in order to fully utilize the textual information in documents. For evaluation, we employ two metrics \textit{Accuracy} and \textit{RMSE}~(Root Mean Square Error), which are calculated using the scripts in \newcite{scikit-learn}\footnote{https://scikit-learn.org/stable/modules/classes.html\#module-sklearn.metrics}. All experiments are conducted on one Nvidia GeForce RTX 3090 GPU.

\subsection{Results}
\begin{table*}[!htb]
  \centering
  \small
  \begin{tabular}{lcccccccccccccccc}
    \toprule
     && \multicolumn{2}{c}{IMDB} && \multicolumn{2}{c}{Yelp-2013}  && \multicolumn{2}{c}{Yelp-2014} \\ 
    
     && Acc. (\%) &  RMSE &&  Acc. (\%) &  RMSE &&  Acc. (\%) & RMSE  \\
     
    \midrule

    \textit{Pre-BERT models} \\
    
    \hspace{0.2cm} {UPNN~\cite{tang-etal-2015-user-product}}  && 43.5 & 1.602 && 59.6  & 0.784 && 60.8  & 0.764  \\
    
    \hspace{0.2cm} {NSC~\cite{chen-etal-2016-neural-sentiment}} && 53.3 & 1.281 && 65.0  & 0.692 && 66.7  & 0.654  \\
    
    \hspace{0.2cm} {UPDMN~\cite{dou-2017-capturing}} && 46.5 & 1.351 && 63.9  & 0.662 && 61.3  & 0.720  \\
    
    \hspace{0.2cm} {CMA~\cite{ma-etal-2017-cascading}}   && 54.0 & 1.191 && 66.3  & 0.677 && 67.6  & 0.637  \\
    
    \hspace{0.2cm} {HCSC~\cite{amplayo-etal-2018-cold}}  && 54.2 & 1.213 && 65.7  & 0.660 && 67.6  & 0.639  \\

    \hspace{0.2cm} {DUPMN~\cite{long-etal-2018-dual}} && 53.9 & 1.279 && 66.2  & 0.667 && 67.6  & 0.639  \\
    
    \hspace{0.2cm} {HUAPA~\cite{aaai-18-jiajun-chen}} && 55.0 & 1.185 && 68.3  & 0.628 && 68.6  & 0.626  \\
    
    \hspace{0.2cm} {RRP-UPM~\cite{cikm19-memory}} && 56.2 & 1.174 && 69.0  & 0.629 && 69.1  & 0.621 \\
    
    \hspace{0.2cm} {CHIM~\cite{amplayo-2019-rethinking}}  && 56.4 & 1.161 && 67.8  & 0.641 && 69.2  & 0.622  \\

   \textit{BERT-based models} \\

    
    
    \hspace{0.2cm} IUPC~\cite{lyu-etal-2020-improving} && 53.8 & 1.151 && 70.5  & 0.589 && 71.2  & 0.592 \\ 

    \hspace{0.2cm} MA-BERT~\cite{zhang-etal-2021-ma-bert} && 57.3 & 1.042 && 70.3  & 0.588 && 71.4  & 0.573 \\ 
    
    \hspace{0.2cm} Ours && \textbf{59.0} & \textbf{1.031} && \textbf{72.1} & \textbf{0.570} && \textbf{72.6} & \textbf{0.563} \\

    \bottomrule
    \end{tabular}%
    \caption{Experimental Results on the test sets of IMDb, Yelp-2013 and Yelp-2014. We report the average results of of five runs of two metrics Accuracy~($\uparrow$) and RMSE~($\downarrow$). The best performance is in bold.}
  \label{tbl-02-main-results}%
\end{table*}

\begin{table*}[!htb]
  \centering
  \small
  \begin{tabular}{lcccccccccccccccc}
    \toprule
     && \multicolumn{2}{c}{IMDB} && \multicolumn{2}{c}{Yelp-2013}  && \multicolumn{2}{c}{Yelp-2014} \\ 
    
     && Acc. (\%) &  RMSE &&  Acc. (\%) &  RMSE &&  Acc. (\%) & RMSE  \\
     
    \midrule

BERT && 50.8 & 1.187 && 67.2  & 0.639 && 67.8  & 0.629 \\
        \hspace{0.1cm} + User-Product Information && 55.4 & 1.129 && 69.1 & 0.617 && 70.7  & 0.610 \\
        \hspace{0.1cm} + Textual Information && 56.9 & 1.089 && 70.1  & 0.593 && 71.9  & 0.563 \\
        \hspace{0.1cm} + User-Product Cross-Context && 59.7 & 1.006 && 70.7 & 0.589 && 72.4  & 0.559 \\

    \bottomrule
    \end{tabular}%
    \caption{Results of ablation studies on the dev sets of IMDb, Yelp-2013 and Yelp-2014.}
  \label{tbl-04-ablation-studies-results}%
\end{table*}
In order to validate the effectiveness of our approach, we first conduct experiments with several PLMs (BERT, SpanBERT and Longformer). Results on the dev sets of IMDb, Yelp-2013 and Yelp-2014 are shown in Table~\ref{tbl-03-more-models-results}. We compare our approach to a vanilla user and product attention baseline
where 1) the user and product representation matrices are randomly initialized and 2) we simply employ multi-head attention between user/product and document representations without the user-product cross-context module. 
Our approach is able to achieve consistent improvements over the baseline  with all PLMs on all three datasets.
For example, our approach gives improvements over the baseline of 4.3 accuracy on IMDb, 1.6 accuracy on Yelp-2013 and 1.7 accuracy on Yelp-2014 for BERT-base. Moreover, our approach can give further improvements for large PLMs such as Longformer-large: improvements of 4.8 accuracy on IMDb, 2.8 accuracy on Yelp-2013 and 2.1 accuracy on Yelp-2014.
The improvements over the baseline are statistically significant ( $p < 0.01$).

We compare our approach 
to previous approaches on the test sets of IMDb, Yelp-2013 and Yelp-2014.  These include  pre-BERT neural baseline models using CNN~\cite{ibm_cnn, kim-2014-convolutional} and LSTM~\cite{yang2016hierarchical_han} -- 
UPNN~\cite{tang-etal-2015-user-product},
NSC~\cite{chen-etal-2016-neural-sentiment},
UPDMN~\cite{dou-2017-capturing},
CMA~\cite{ma-etal-2017-cascading},
HCSC~\cite{amplayo-etal-2018-cold},
DUPMN~\cite{long-etal-2018-dual},
HUAPA~\cite{aaai-18-jiajun-chen},
RRP-UPM~\cite{cikm19-memory},
CHIM~\cite{amplayo-2019-rethinking} -- and two state-of-the-art models based on BERT including IUPC~\cite{lyu-etal-2020-improving} and MA-BERT~\cite{zhang-etal-2021-ma-bert}. We use BERT-base for a fair comparison with IUPC and MA-BERT, which both use BERT-base. 
The results are shown in Table~\ref{tbl-02-main-results}.
Our model obtains the best performance at both accuracy and RMSE on IMDb, Yelp-2013 and Yelp-2014. Specifically, our model achieves absolute improvements in accuracy of 1.7, 1.6 and 1.2 on IMDb, Yelp-2013 and Yelp-2013 respectively compared to previous state-of-the-art results. As for RMSE, which indicates how \textit{close} the predicted labels are to ground-truth labels, our models outperforms earlier state-of-the-art models on RMSE by 0.011 on IMDb, 0.018 on Yelp-2013 and 0.010 on Yelp-2014. 

\begin{table*}[!htb]
  \centering
  \begin{tabular}{lccccccc}
\toprule

\multirow{3}{6em}{\centering IMDb} & Max Length  & 128 & 256  & 384 & 512  & 1024 & 2048 \\
    & Truncated Examples~(\%) & 96.3 & 68.7 & 46.5 & 30.8 & 6.3 & 0 \\
        & Accuracy~(\%) &  33.9 &  37.2 & 45.0 & 54.3 & 58.4 & 58.9  \\

    \midrule
    \multirow{3}{6em}{\centering Yelp-2013} & Max Length  & 128 & 256  & 384 & 512  & 1024 & 2048 \\
    & Truncated Examples~(\%) & 63.7 & 29.3 & 13.1 & 5.6 & 0.3 & 0 \\
    & Accuracy~(\%) &  63.1 &  66.6 & 68.1 & 68.6 & 69.4 & 69.4  \\
    \bottomrule
\end{tabular}%
    \caption{Results of Longformer under different maximum sequence length on the dev sets of IMDb and Yelp-2013. The truncated examples are the percentage of examples that exceed the corresponding max sequence length.}.
  \label{tbl-06-longformer-len}%
\end{table*}

\begin{figure}
    \centering
    \includegraphics[width=\linewidth]{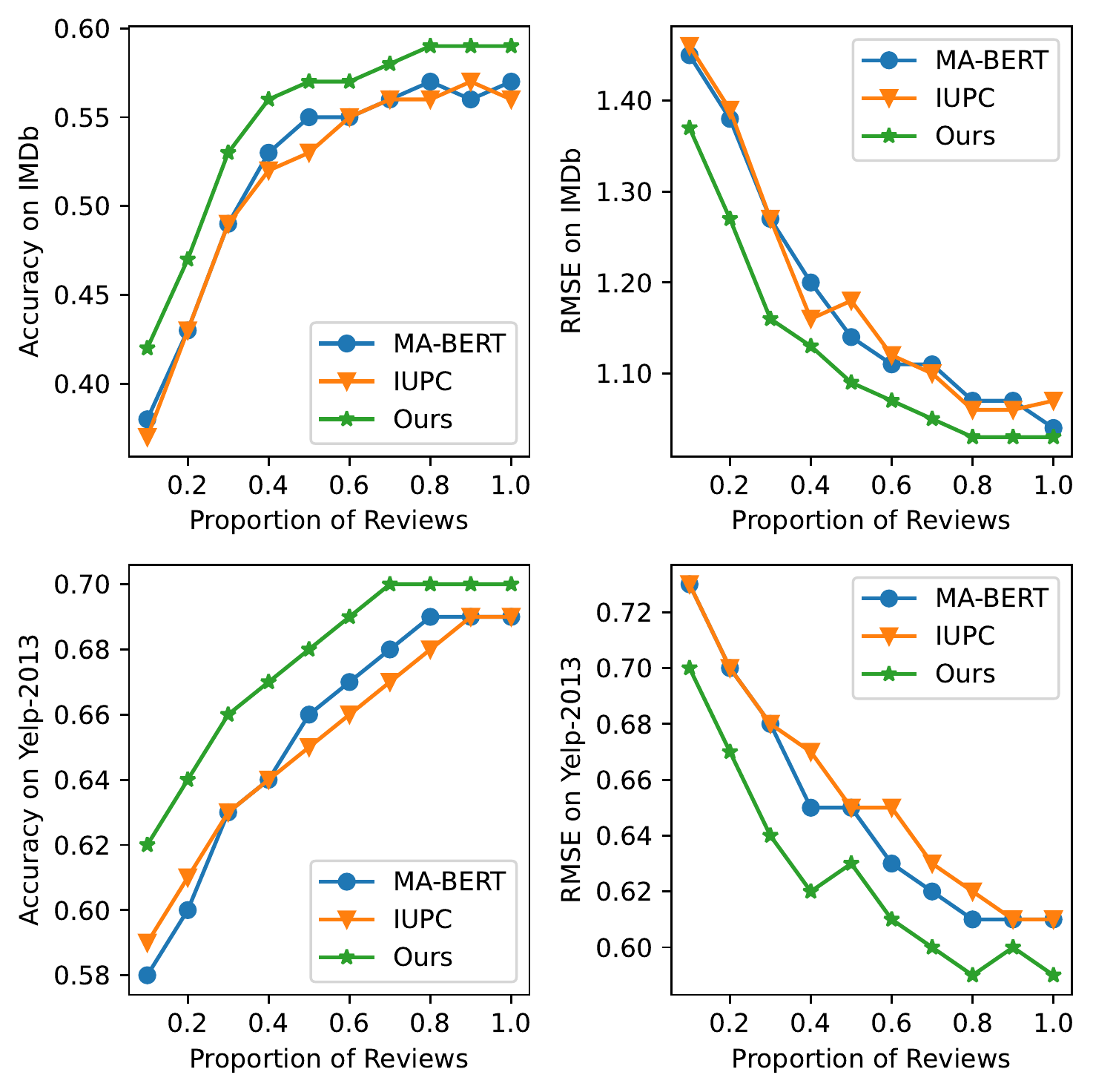}
    \caption{Experimental results of IUPC, MA-BERT and our approach under different proportion of reviews from 10\% to 100\% on the dev sets of IMDb and Yelp-2013.}
    \label{fig:user_ratio}
\end{figure}
\subsection{Ablation Studies} 
Results of an ablation analysis are shown in Table~\ref{tbl-04-ablation-studies-results}.
The first row results are from a BERT model without user and product information.
The next three rows correspond to
\begin{enumerate}
\setlength\itemsep{-0.5em}
    \item \textit{User-Product Information}, where we use the same method in the baseline vanilla attention model in Table~\ref{tbl-03-more-models-results} to inject user-product information
    \item \textit{Textual Information}, our proposed approach of using historical reviews to initialize user and product representations.
    \item \textit{User-Product Cross-Context}, our proposed module incorporating the associations between users and products.
\end{enumerate}
The results show, firstly, that user and product information is highly useful for sentiment classification, and, secondly, that both textual information of historical reviews and user-product cross-context can improve sentiment classification. For example, \textit{Textual Information} gives \char`\~ 1 accuracy improvement on three datasets while giving \char`\~0.04 RMSE improvement on IMDb and Yelp-2014, \char`\~ 0.02 RMSE improvement on Yelp-2013. \textit{User-Product Cross-Context} achieves large improvements on IMDb of 2.8 accuracy compared to the improvements on Yelp-2013 and Yelp-2014 of 0.6 and 0.5 accuracy respectively.

\subsection{Performance under Varying Amount of Reviews to User/Product}

\begin{figure}
    \centering
    \includegraphics[width=\linewidth]{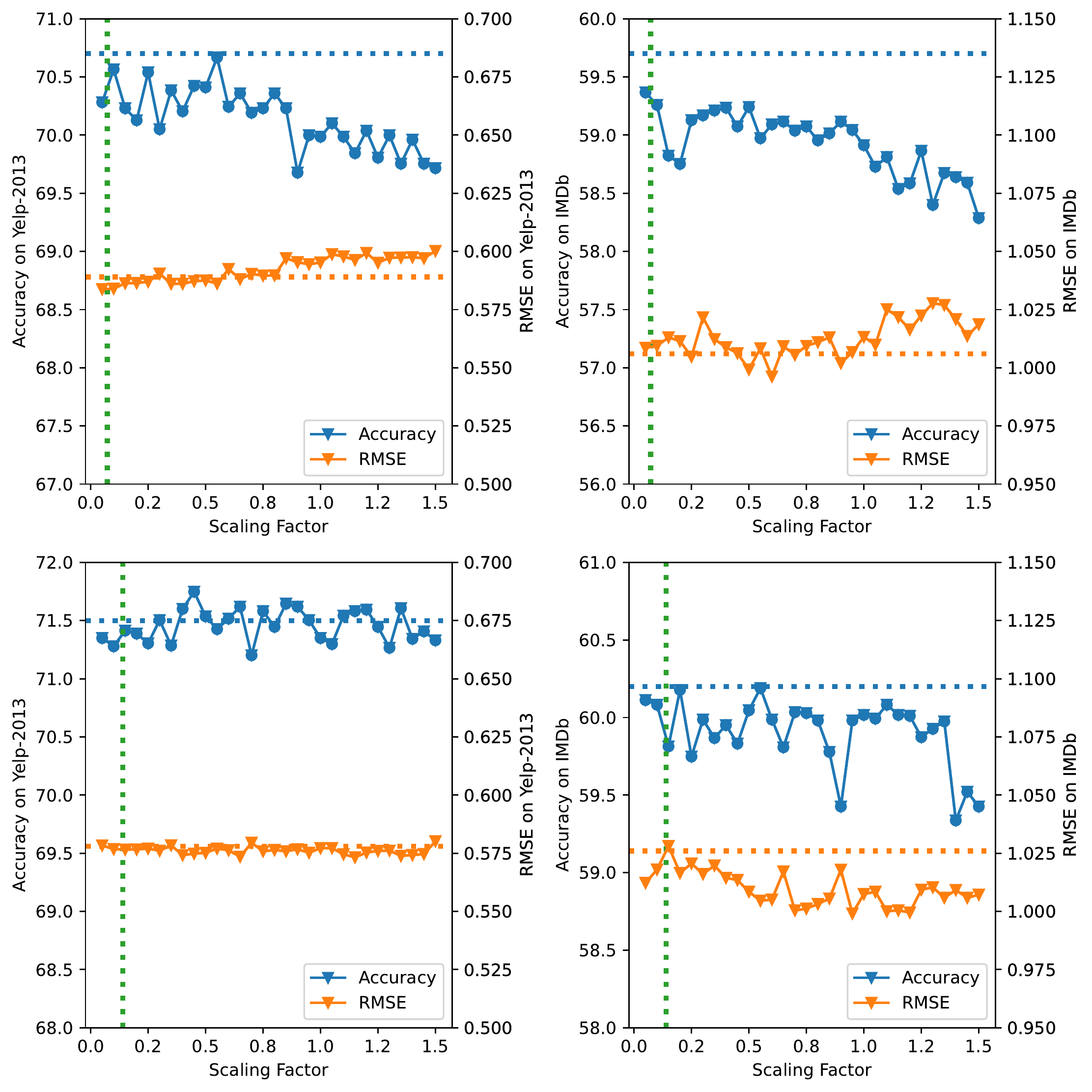}
    \caption{Effect of varying the scaling factor for the User and Product Matrices on the dev sets of Yelp-2013~(left) and IMDb~(right). We include results of BERT-base~(top) and SpanBERT-base~(bottom). The left and right y-axis in each subplot represent \textit{Accuracy} and \textit{RMSE} respectively. The x-axis represents the scaling factor. The vertical green dashed line is the scaling factor from the Frobenius norm heuristic. The two horizontal dashed lines~(blue and orange) are the accuracy and RMSE produced by the Frobenius norm heuristic respectively.}
    \label{fig:effect_of_scaling_factor}
\end{figure}

We investigate model performance with different amounts of reviews belonging to the same user/product. 
We randomly sample a proportion of each user's reviews (from 10\% to 100\%). 
Then we use the sampled training data, where each user only has part of their total reviews (e.g. 10\%), to train sentiment classification models. 
We conduct experiments on Yelp-2013 using IUPC~\cite{lyu-etal-2020-improving}, MA-BERT~\cite{zhang-etal-2021-ma-bert} and our approach. The results are shown in Figure~\ref{fig:user_ratio}, where the x-axis represents the proportion of reviews that we used in experiments. 
When the proportion of reviews lie between 10\% and 50\%, our approach obtains superior performance compared to MA-BERT and IUPC while the performance gain decreases when users have more reviews. The results show the advantage of our approach under a low-review scenario for users. 

\begin{table*}[!htb]
  \centering
  \small
  \resizebox{\textwidth}{!}{
  \begin{tabular}{p{8cm}p{1.35cm}p{1.35cm}p{1.35cm}p{1.35cm}}
    \toprule
     Review & Vanilla BERT & IUPC & MA-BERT & Ours \\
     
    \midrule
    \textit{Took travis here for one of our first dates and just love cibo. It 's situated in a home from 1913 and has colored lights wrapped all around the trees. You can either sit inside or on the gorgeous patio. Brick oven pizza and cheese plates offered here and it 's definitely a place for a cool date.} (VP) & VP~(\cmark)  & VP~(\cmark)  & VP~(\cmark) & VP~(\cmark)\\
\hline
\textit{a great sushi bar owned and operated by maggie and toshi who are both japanese. their product is always consistent and they always have a few good specials. service is great and the staff is very friendly and cheerful. value is really good particularly within their happy hour menu. our kids love it and they are always spoiled rotten by maggie and toshi so it is their favorite place. lastly we did a sake tasting there a few weeks ago and really had a great time. we all sat family style int he middle of the restaurant and got to experience some really interesting rice wines. we had a blast. great place}  (P) & VP~(\xmark)  & P~(\cmark)  & P~(\cmark) & P~(\cmark)  \\

\hline
\textit{well , i was disappointed. i was expecting this one to be a jazzed up container store. but ... it was just average. i used to visit container store in houston near the galleria. it has a nice selection of things. people are always ready to help etc.. but , this one has an aloof sort of customer service crowd. they say nice things about your kid but do not offer to help. hmm ... i have seen similar things they were selling at ikea. the quality did seem a little better than ikea but if you are buying a laundry room shelf for your laundry detergent ... who the hell cares. its a shelf !  does n't matter if it has 15 coats of paint on the metal or 2 coats. i found one of those sistema lunch boxes that i have been looking for over here and it was on sale. will i go back ?  probably not. too far out for me , plus i like ikea better} (Ne) & VN~(\xmark)  & N~(\xmark)  & VN~(\xmark) & Ne~(\cmark) \\


\hline
\textit{Unfortunately tonight was the last night this location was open. The only two locations left in the valley are desert ridge and arrowhead. Please support them.} (VP) & Ne~(\xmark)  & N~(\xmark)  & VN~(\xmark) & N~(\xmark) \\ 

    \bottomrule
    \end{tabular}%
  }

    \caption{Example reviews from the dev sets of Yelp-2013 and the corresponding predictions of each model. Very Negative~(VN), Negative~(N), Neutral~(Ne), Positive~(P), Very Positive~(VP).}
  \label{tbl-05-cases}%
\end{table*}
\subsection{Scaling Factor for User/Product Matrix}
We conduct experiments with different scaling factor (see Equations~\ref{user_scaling} and~\ref{product_scaling}) on the dev sets of Yelp-2013 and IMDb using BERT-base. In this experiment, we apply the same scaling factor to both user and product matrix. The results are shown in Figure~\ref{fig:effect_of_scaling_factor}, where we use scaling factor ranging from 0.05 to 1.5 with intervals of 0.05. The results 
show that our proposed scaling factor~(green dashed lines in Figure~\ref{fig:effect_of_scaling_factor}) based on Frobenius norm can yield competitive performance: best accuracy according to the blue dashed line. Although the RMSE of Frobenius norm heuristic is not always the optimal, it is still a relatively lower RMSE compared to most of the other scaling factors~(except the RMSE of SpanBERT-base on IMDb). Moreover, the Frobenius norm heuristic can reduce the efforts needed to tune the scaling factor, since the optimal scaling factor is varying for different models on different data, whereas the Frobenius norm heuristic is able to consistently provide a competitive dynamic scaling factor.

\subsection{Effect of Maximum Sequence Length}
Document length can make document-level sentiment classification more challenging, especially for fine-grained classification, which requires model to capture the subtle expression of sentiment polarity in documents. However, PLMs often have a fixed maximum sequence length (usually 512 WordPiece~\cite{wu2016google_nmt_system} tokens). 
A commonly used method for dealing with this constraint is to only keep the first 512 tokens for documents longer than the maximum length. This has been shown, however, not to be the best strategy~\cite{sun2019fine_bert_qiu}, because the expression of sentiment polarity could be towards the end of a document. 
Therefore, in order to investigate the importance to sentiment polarity prediction of text in the tail end of a long review, we conduct experiments on the dev sets of IMDb and Yelp-2013 using Longformer-base~\cite{beltagy2020longformer}. We adopt various maximum sequence lengths, from 64 to up the 2048 tokens handled by Longformer. In order to purely focus on review texts, we do not include user/product information in this experiment.

The results are shown in Table~\ref{tbl-06-longformer-len}. When reviews longer than the maximum length are truncated, the performance of sentiment classification is substantially reduced. For example, in IMDb, when the maximum length is set to 128 and 256, 96.3\% and 68.7\% examples are truncated and the accuracy drops \char`\~ 40\% compared to the best performance. However, the effect  is lower for Yelp-2013. For example, when 63.7\% and 29.3\% examples are truncated, the accuracy only drops \char`\~ 10\% and \char`\~ 5\% compared to the best accuracy.
This is not surprising given the shorter review length of  Yelp versus IMDb reviews (see Table~\ref{data_statistics_train}). 

\subsection{Examples}
Some cases sampled from the dev set of Yelp-2013 and corresponding predictions from Vanilla BERT w/o user and product information, IUPC~\cite{lyu-etal-2020-improving}, MA-BERT~\cite{zhang-etal-2021-ma-bert} and our model are shown in Table~\ref{tbl-05-cases}. 
\paragraph{Example 1}This is a straightforward positive review since it clearly conveys the satisfaction towards the restaurant. Thus all models make the correct prediction.
\paragraph{Example 2} This is similar to the first example in narrative style, but the ground-truth sentiment label is Positive rather than Very Positive since this user tends not to give very high ratings. 
This example shows the importance of user information. 
\paragraph{Example 3} 
This review conveys a very negative attitude. However, the author tends not to give very poor ratings 
plus the reviews this store received are not bad. With both user and product information, our model makes the correct prediction of Neutral. 
\paragraph{Example 4} 
 All models, regardless of whether they use user and product information, predict Neutral or Negative  while in fact the review label is Very Positive.
 This is a difficult example where the sentiment is subtly expressed. 

\section{Conclusion and Future Work}
In order to make the best use of user and product information in document-level sentiment classification, we propose a text-driven approach: 1) explicitly utilizing historical reviews to initialize user and product representations 2) modeling associations between users and products with an additional user-product cross-context module. The experiments conducted on three benchmark datasets including IMDb, Yelp-2013 and Yelp-2014 demonstrate that our approach substantially outperforms previous state-of-the-art approaches and is effective for several PLMs. We also show that our method obtains superior performance when the amount of reviews for a user/product is limited. For future work, we aim to apply our approach to more tasks where there is a need to learn representations for various types of attributes.


\section*{Acknowledgements}
This work was funded by Science Foundation Ireland through the SFI Centre for Research Training in Machine Learning (18/CRT/6183).

\bibliography{anthology,custom}
\bibliographystyle{acl_natbib}

\appendix

\section{Appendix}
\label{sec:appendix}

\subsection{Hyperparameters}

We show the Learning Rate and Batch Size used to train our models on all datasets in Table~\ref{tbl-07-hyperparameters}.
\begin{table*}[!htb]
  \centering
  \begin{tabular}{lcccccccccccccccc}
    \toprule
     && \multicolumn{2}{c}{IMDB} && \multicolumn{2}{c}{Yelp-2013}  && \multicolumn{2}{c}{Yelp-2014} \\ 
    
     && BS &  LR &&  BS &  LR &&  BS & LR  \\
     
    \midrule

         BERT-base && 16 & 6e-5 && 16 & 6e-5 && 16 & 6e-5 \\
        
        
         BERT-large && 8 & 3e-5 && 8 & 3e-5 && 8 & 3e-5 \\
        
        
         SpanBERT-base && 16 & 6e-5 && 16 & 6e-5 && 16 & 6e-5 \\
        
        
         SpanBERT-large && 8 & 3e-5 && 8 & 3e-5 && 8 & 3e-5 \\

         Longformer-base && 16 & 3e-5 && 16 & 3e-5 && 16 & 3e-5 \\
        
         Longformer-large && 4 & 2e-5 && 4 & 3e-5 && 4 & 3e-5 \\
    
    \bottomrule
    \end{tabular}%
    \caption{The hyperparameters used to fine-tune all models on all datasets including Learning Rate~(LR) and Batch Size~(BS).}
  \label{tbl-07-hyperparameters}%
\end{table*}




\end{document}